\documentclass[runningheads]{llncs}
\usepackage{graphicx}
\usepackage{booktabs}
\usepackage{multirow}
\usepackage[normalem]{ulem}
\usepackage{array}
\usepackage{ragged2e}

\usepackage{ulem}
\usepackage{xcolor}
\usepackage{hyperref}

\begin{document}
\title{DETDet: Dual Ensemble Teeth Detection}
%
%
\author{Kyoungyeon Choi\inst{1,2}\orcidID{0009-0004-1123-3237} \and
Jaewon Shin\inst{1,2}\orcidID{0009-0006-9044-241X} \and
Eunyi Lyou\inst{1,3}\orcidID{0009-0004-1378-9546}}
\authorrunning{Choi. Author et al.}
%
\institute{
Evident  Co., Ltd., Seoul, Korea\\
\email{\{besteverchoi, eunyi.lyou\}@gmail.com}\\ \and
School of Dentistry, Seoul National University \and
Graduate School of Data Science, Seoul National University
}
\maketitle              
\begin{abstract}
The field of dentistry is in the era of digital transformation. Particularly, artificial intelligence is anticipated to play a significant role in digital dentistry. AI holds the potential to significantly assist dental practitioners and elevate diagnostic accuracy. In alignment with this vision, the 2023 DENTEX challenge aims to enhance the performance of dental panoramic X-ray diagnosis and enumeration through technological advancement. In response, we introduce \textbf{DETDet}, a Dual Ensemble Teeth Detection network. DETDet encompasses two distinct modules dedicated to enumeration and diagnosis. Leveraging the advantages of teeth mask data, we employ Mask-RCNN for the enumeration module. For the diagnosis module, we adopt an ensemble model comprising DiffusionDet and DINO. To further enhance precision scores, we integrate a complementary module to harness the potential of unlabeled data. The code for our approach will be made accessible at \href{https://github.com/Bestever-choi/Evident}{https://github.com/Bestever-choi/Evident}

\end{abstract}

\section{Method}
\paragraph{Data preprocessing}
All of the training X-ray images were normalized using a mean of 0.5 and a standard deviation of 0.1. We employed random resizing in the enumeration module and included a random horizontal flip in the diagnosis module. For model validation and testing, we split the enumeration dataset having 634 panoramic dental X-rays into 534 images for training, 50 images for validation, and 50 images for test. Similarly, diagnosis datasest containing 705 images were divided into 605 training data, along with 50 instances each for validation and testing.

\begin{figure}
\includegraphics[width=\textwidth]{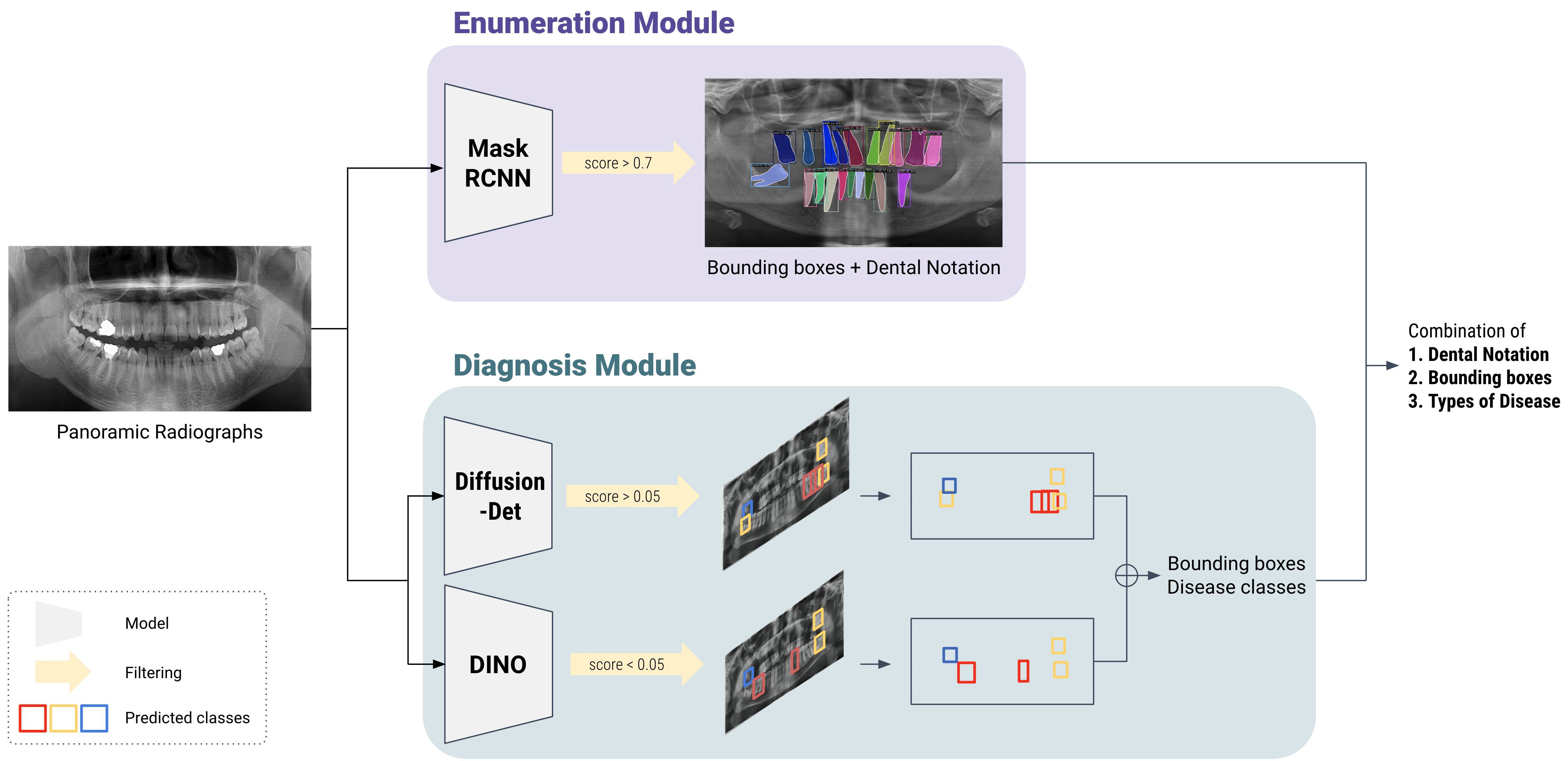}
\caption{Model Architecture of DETDet.} \label{fig:overview}
\end{figure}

\paragraph{Overview} In our proposed method, illustrated in Fig. \ref{fig:overview}, we input a panoramic radiograph into the \textbf{enumeration module}, utilizing mask RCNN to predict dental bounding boxes and notations. Simultaneously, the same radiograph is fed into the \textbf{diagnosis module}, an ensemble of DiffusionDet and DINO, which diagnoses diseases and predicts bounding boxes. These outputs are integrated to yield comprehensive dental notations and bounding boxes for teeth affected by diseases.

\subsection{Enumeration module}
Given the 705 training images for quadrant-enumeration-disease data, we observed that there are only a few labels of enumeration in the disease data. To address this problem of small number of labels, HierarchicalDet[1] applied weight transfer to learn enumeration bounding boxes from the 634 enumeration-only data. Instead, we used the instance segmentation method to increase the mean average precision (mAP). Generally, instance segmentation yields better mean average precision (mAP) than detection algorithms, as the segmentation network learns additional mask information\cite{ref_article2}, enabling to understand finer-grained signals in images. Specifically, we trained Mask-RCNN with the SwinT backbone on the 634 enumeration-only data. 
As using Mask-RCNN with SwinT backbone outperformed SOTA methods(HTC\cite{ref_article3} and Cascade Mask-RCNN\cite{ref_article4}) in the given enumeration data, we adopted the former to predict quadrant and numerical index of each tooth, along with their corresponding bounding boxes. Finally, we only considered the bounding box predictions with scores higher than 0.7 to exclude low-scoring enumeration outputs.

\subsection{Ensembling diagnosis module}
In our preliminary experiment, two state-of-the-art methods, DiffusionDet[5] and DINO[7], a DETR variant with improved denoising anchor boxes, were evaluated on diagnosis dataset. While DiffusionDet yields a higher Average Precision than other object detection algorithms, DINO leads to a higher Average Recall than any other detection methods. Empirically, the large number of low-score DINO detections, approximately 3000 detections per image, contributes to the high Average Recall, whereas DiffusionDet predicts a substantial number of high-scoring true positives leading to the high Average Precision. Therefore, we employ an ensemble approach that combines both methods to achieve high scores in both Average Precision and Average Recall. Specifically, we utilize DiffusionDet predictions with bounding box scores higher than 0.05 and DINO predictions with scores lower than 0.05 to increase the Average Recall while maintaining precision. Additionally, it is noteworthy that, according to our test settings, DiffusionDet with the ResNet50 backbone outperformed SwinT. Hence, we use the ResNet50 backbone for both DINO and DiffusionDet in the diagnosis module.

\subsection{Dual stage integration}
In order to integrate the enumeration and diagnosis modules, we introduce a method called \textit{closest bounding box center matching}. This method operates in two stages: it first predicts the enumeration bounding boxes and then matches these boxes to the nearest diagnosis bounding boxes. Through this integration, we are able to derive category IDs for all three elements: quadrant, enumeration, and disease. The combined bounding box score is computed by multiplying the scores from the enumeration and diagnosis modules. In summary, DETDet integrates enumeration and diagnosis through a dual-stage approach, ensembling two models to achieve high precision and recall.

\subsection{Complementary module}

\begin{figure}
\includegraphics[width=\textwidth]{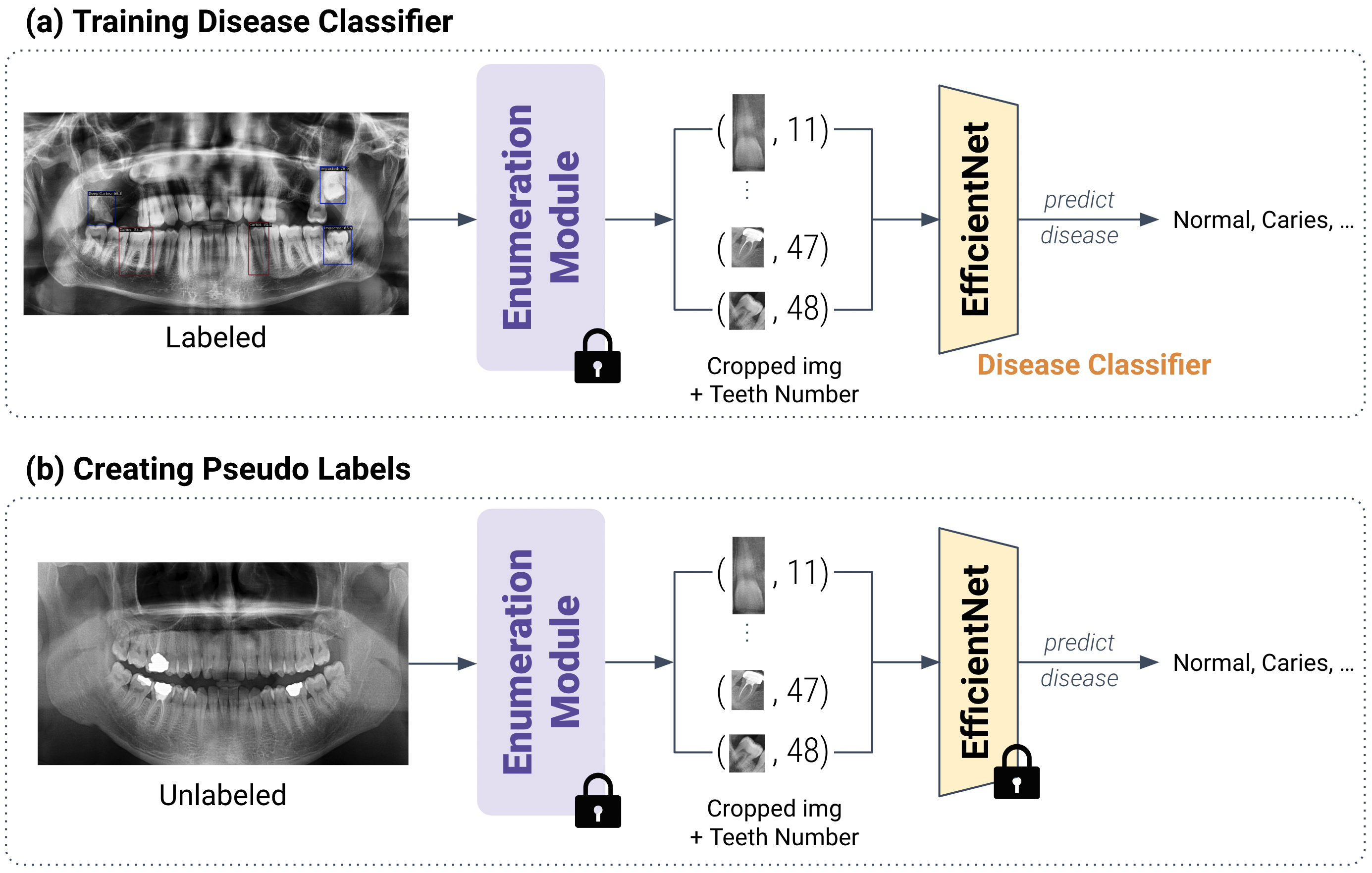}
\caption{Training process of the complementary module} \label{fig:complementary}
\end{figure}
Furthermore, the dual-stage integration offers the advantage of explicitly utilizing the 1571 unlabeled data dealing with the insufficient amount of original training data. The training data suffers from significant imbalance, as there is a notable disparity in the amount of caries data compared to other diseases. In particular, the dataset contains a scarcity of Periapical Lesion data. To address this data imbalance, we introduce a complementary module, as depicted in Fig. \ref{fig:complementary}, which is designed to leverage unlabeled data for increasing the number of train samples. The complementary module is trained using a pseudo-labeling approach. Specifically, we segment all the dental images using the enumeration module and assign each tooth to its corresponding category. Subsequently, an EfficientNetB4[6] classifier is trained to classify teeth into four categories: normal, caries, deep caries, impacted, and periapical lesion. We apply this classifier to the unlabeled data, assigning pseudo labels to all teeth. Through pseudo labeling, we are able to augment the periapical lesion data, thereby achieving a more balanced dataset overall. While the complementary module generates fewer predictions compared to the diagnosis module, it effectively compensates for the predictions that the diagnosis module may miss.

\section{Results}
\subsection{Enumeration results}
The enumeration module, trained using Mask-RCNN with the SwinT backbone, is evaluated on our set of 50 test images. The outcomes are detailed in Table \ref{table:enum}. The AP50 score stands at 0.987, indicating that the module adeptly classifies and detects teeth with a high level of precision. Nevertheless, the AP75 score is relatively lower. This discrepancy can be attributed to variations in the ground truth bounding boxes, as they were labeled by different individuals, resulting in potential differences in exact bounding box coordinates across all images.



\begin{table}
\centering
\caption{Results of Enumeration Module}
\label{table:enum}
\begin{tabular}{>{\centering\hspace{0pt}}m{0.1\linewidth}>{\centering\hspace{0pt}}m{0.2\linewidth}>{\centering\arraybackslash\hspace{0pt}}m{0.2\linewidth}} 
\toprule
\multirow{2}{0.1\linewidth}{\hspace{0pt}\Centering{}} & \multicolumn{2}{>{\Centering\hspace{0pt}}m{0.4\linewidth}}{Task} \\ 
\cmidrule(lr){2-2}\cmidrule(lr){3-3}
 & Bounding box & Segmentation \\ 
\midrule
mAP & 0.589 & 0.548 \\
AP50 & 0.987 & 0.976 \\
AP75 & 0.636 & 0.552 \\
AR & 0.665 & 0.615 \\
\bottomrule
\end{tabular}
\end{table}

\subsection{Ensemble diagnosis results}

The diagnosis module, trained using DiffusionDet and DINO with a ResNet50 backbone, is evaluated on our set of 50 test images. The outcomes are outlined in Table \ref{table:diag}. The mAP, AP50, and AP75 scores are notably higher in DiffusionDet compared to DINO. However, the AR score is significantly superior in DINO, owing to its output of 32\% more detections than DiffusionDet. Consequently, we combine these two models in an ensemble approach to capitalize on their strengths. The ensemble module selects the DiffusionDet model when the prediction score exceeds a certain threshold, otherwise it selects DINO. After conducting experimentation on the test data, we set the threshold at 0.05. This strategic choice yields a slight increase in mAP and AP50, along with a substantial improvement in AR. The results indicate a notable enhancement in overall performance.

\begin{table}
\centering
\caption{Results of Diagnosis Module}
\label{table:diag}
\begin{tabular}{>{\hspace{0pt}}m{0.2\linewidth}>{\centering\hspace{0pt}}m{0.12\linewidth}>{\centering\hspace{0pt}}m{0.12\linewidth}>{\centering\hspace{0pt}}m{0.12\linewidth}>{\centering\arraybackslash\hspace{0pt}}m{0.12\linewidth}} 
\toprule
\multicolumn{1}{>{\Centering\hspace{0pt}}m{0.2\linewidth}}{\multirow{2}{0.2\linewidth}{\hspace{0pt}\Centering{}Model}} & \multicolumn{4}{>{\Centering\hspace{0pt}}m{0.48\linewidth}}{Metric} \\ 
\cmidrule(l){2-5}
\multicolumn{1}{>{\Centering\hspace{0pt}}m{0.2\linewidth}}{} & mAP & AP50 & AP75 & AR \\ 
\midrule
DiffusionDet & 0.373 & \textbf{0.614} & 0.436 & 0.634 \\
DINO & 0.282 & 0.438 & 0.323 & 0.705 \\
Ensemble & \textbf{0.380} & 0.610 & \textbf{0.451} & \textbf{0.715} \\
\bottomrule
\end{tabular}
\end{table}

\subsection{Dual stage integration results}
The outcomes of the dual-stage integration are outlined in Table \ref{table:dual}. We employed the closest bounding box matching method to integrate the enumeration and diagnosis modules, enabling the prediction of category IDs 1, 2, and 3.

\begin{table}
\centering
\caption{Results of Dual-Stage Integration }
\label{table:dual}
\begin{tabular}{>{\hspace{0pt}}m{0.2\linewidth}>{\centering\hspace{0pt}}m{0.12\linewidth}>{\centering\hspace{0pt}}m{0.12\linewidth}>{\centering\hspace{0pt}}m{0.12\linewidth}>{\centering\arraybackslash\hspace{0pt}}m{0.12\linewidth}} 
\toprule
\multicolumn{1}{>{\Centering\hspace{0pt}}m{0.2\linewidth}}{\multirow{2}{0.2\linewidth}{\hspace{0pt}\Centering{}}} & \multicolumn{4}{>{\Centering\hspace{0pt}}m{0.48\linewidth}}{Metric} \\ 
\cmidrule(l){2-5}
\multicolumn{1}{>{\Centering\hspace{0pt}}m{0.2\linewidth}}{} & mAP & AP50 & AP75 & AR \\ 
\midrule
Quadrant & 0.404 & 0.638 & 0.479 & 0.749 \\
Enumeration & 0.749 & 0.365 & 0.246 & 0.643 \\
Disease & 0.380 & 0.610 & 0.451 & 0.715 \\
\bottomrule
\end{tabular}
\end{table}

\subsection{Complementary module results}
The inclusion of the complementary module yields an overall enhancement in the metrics. The outcomes are consolidated in Table \ref{table:complementary}. Notably, there is a substantial increase in the AP50 scores for both disease and enumeration. The complementary module is trained using an expanded dataset, which encompasses unlabeled data. In particular, we augmented the periapical lesion and deep caries data by a factor of two, thereby mitigating the training data's imbalance. Consequently, the complementary module contributes supplementary predictions to the dual-stage integration outputs, resulting in improved precision and recall.

\begin{table}
\centering
\caption{Results of Complementary Module}
\label{table:complementary}
\begin{tabular}{>{\hspace{0pt}}m{0.2\linewidth}>{\centering\hspace{0pt}}m{0.12\linewidth}>{\centering\hspace{0pt}}m{0.12\linewidth}>{\centering\hspace{0pt}}m{0.12\linewidth}>{\centering\arraybackslash\hspace{0pt}}m{0.12\linewidth}} 
\toprule
\multicolumn{1}{>{\Centering\hspace{0pt}}m{0.2\linewidth}}{\multirow{2}{0.2\linewidth}{\hspace{0pt}\Centering{}}} & \multicolumn{4}{>{\Centering\hspace{0pt}}m{0.48\linewidth}}{Metric} \\ 
\cmidrule(l){2-5}
\multicolumn{1}{>{\Centering\hspace{0pt}}m{0.2\linewidth}}{} & mAP & AP50 & AP75 & AR \\ 
\midrule
Quadrant & 0.410 & 0.633 & 0.494 & 0.768 \\
Enumeration & 0.292 & 0.463 & 0.344 & 0.672 \\
Disease & 0.402 & 0.668 & 0.441 & 0.744 \\
\bottomrule
\end{tabular}
\end{table}

%
%
%
%

\end{document}